# Enhanced CNN for image denoising

*Chunwei Tian[1,2], Yong Xu[1,2] ✉, Lunke Fei[3], Junqian Wang[1,2], Jie Wen[1,2], Nan Luo[4]*

*[1]Bio-Computing Research Center, Harbin Institute of Technology, Shenzhen, Shenzhen 518055, People's Republic of China*
*[2]Shenzhen Medical Biometrics Perception and Analysis Engineering Laboratory, Harbin Institute of Technology, Shenzhen, Shenzhen 518055, People's Republic of China*
*[3]School of Computer Science and Technology, Guangdong University of Technology, Guangzhou 510006, People's Republic of China*
*[4]Institute of Automation Heilongjiang Academy of Sciences, Harbin 150090, People's Republic of China*
✉ *E-mail: yongxu@ymail.com*

**Abstract**: Owing to the flexible architectures of deep convolutional neural networks (CNNs) are successfully used for image denoising. However, they suffer from the following drawbacks: (i) deep network architecture is very difficult to train. (ii) Deeper networks face the challenge of performance saturation. In this study, the authors propose a novel method called enhanced convolutional neural denoising network (ECNDNet). Specifically, they use residual learning and batch normalisation techniques to address the problem of training difficulties and accelerate the convergence of the network. In addition, dilated convolutions are used in the proposed network to enlarge the context information and reduce the computational cost. Extensive experiments demonstrate that the ECNDNet outperforms the state-of-the-art methods for image denoising.

## 1 Introduction

Image denoising is a classical technique of image restoration and has been successful in many fields such as pathological analysis and human entertainment [1, 2]. The degradation model is widely used in denoising problem to recover clear image, which is expressed as $y = x + \mu$, where $x$ is a clean image, $y$ is a noisy image and $\mu$ is the additive Gaussian noise with standard deviation $\sigma$. According to the Bayesian theory, it is known that the prior is very important for image denoising [3]. For example, wavelet transformation with a prior of Markov random field is used to suppress noise [4]. Combing the self-similarities and sparse representation can improve the performance and reduce the storage for image denoising [5]. Block-matching and 3D filtering (BM3D) converts 2D image data into 3D data arrays and uses the sparse method to deal with the obtained 3D data arrays to remove noise [6]. Enforcing the gradient histogram of the noisy image is approximate to the theoretical gradient histogram of the clean image for image denoising [7]. In addition, Nonlocally centralised sparse representation (NCSR) [8], gradient methods [9, 10], total-variation methods [11, 12] and weight nuclear norm minimisation (WNNM) [13] are also very effective for image denoising.

Although the above methods have obtained great performance for denoising task, they still face the following problems [3]: (i) they need to set manually the parameters to obtain the optimal results. (ii) They use complex optimisation to improve the performance, which increases the computational cost.

Owing to the flexible connection fashion of the deep network architecture and strong learning ability, deep learning techniques have become the most effective methods to address the above problems for image denoising. Specifically, deep convolutional neural networks (CNNs) have attracted more attention in image denoising [14]. For example, CNN uses residual learning method to improve the performance in image denoising [3]. It first uses a model to deal with multiple restoration tasks such as image denoising, image super-resolution and image deblocking. The fusion of CNN and characteristics of denoisng task is useful to remove unknown [15]. Combining CNN and nature of images is very effective to obtain a clean image. For example, CNN utilises non-local similarity to deal with colour noisy images [16]. Discriminative learning methods embedded into optimisation method obtain great performance for real noisy images [17]. CNN consolidated unsupervised learning is a good choice for image restoration [18]. Using the principle of enhanced signal-to-design novel network architecture is also very popular to recover image [19]. Integrating spatial domain into CNN can better filter noise [20]. The combination of traditional denoising methods and CNN such as BMCNN is very competent to separate noise from noisy image [21]. The fusion of multiple features is very beneficial for image denoising [22]. Deep CNN has good visual effects on multiplicative noises [23]. Deep CNN is a good tool for medical image denoising [24, 25]. The recently proposed deep cascade convolutional residual denoising network (DCCRDN) repeatedly uses concatenate operations to train the models for image denoising [26]. Although the above deep network methods have obtained great performance for denoising tasks, most of these methods suffer from the drawbacks of vanishing or exploding gradients when the network architecture is very deep. In addition, the above methods sacrifice the computational cost to improve the performance. For example, they apply multiple concanation operations to train the denoising model.

In this paper, we propose a novel network referred to as enhanced convolutional neural denoising network (ECNDNet). ECNDNet utilises residual learning technique [27] to prevent vanishing and exploding gradient problems. Moreover, batch normalisation (BN) [28] is used to accelerate the convergence of the trained model and make the network easy to train. To decrease the computational burden, we use dilated convolution [29] to capture more context information. Extensive experiments demonstrate that our proposed ECNDNet method outperforms the popular image denoising methods such as fast and flexible denoising net (FFDNet) [15], image restoration CNN (IRCNN) [17] and BM3D [6].

The main contributions of this work are summarised as follows:

(i) The depth of the proposed ECNDNet is only set to 17 layers, which can effectively reduce the computational cost.
(ii) ECNDNet uses residual learning mechanism to prevent vanishing and exploding gradient problems. Besides, it utilises BN technique to normalise data and improve the efficiency of the training model.

(iii) ECNDNet uses dilated convolutions to enlarge the receptive field and improve the performance.

The remaining of this paper is organised as follows. Section 2 presents related work of the proposed method. Section 3 provides the proposed method. Section 4 shows the extensive experimental results of this paper. Section 5 offers the conclusion.

## 2 Related work

### 2.1 BN and residual learning

One of the reasons for CNN's success is its end-to-end connection. The end-to-end connection architecture of CNN generally includes initial parameter [30], gradient optimisation methods [31, 32] and rectified linear unit (ReLU) [33]. Although the general network architecture has obtained good performance, they face vanishing/exploding gradient problems and have difficulty in training deep networks. In this paper, we use BN and residual learning to address the above problems. The detailed information about BN [27] and residual learning [28] are explained as follows: the distribution of sample data is changed after it passes the convolution layer. This phenomenon is called internal covariate shift problem. This problem can be addressed by BN technique. That is, first, BN normalises the training data in every batch. Then, it uses scale and shift operations to recover the distribution of training data. The above two important parameters of BN are updated when the trained network is back propagation. BN is set before the activation function of each layer. BN enjoys the following merits: (i) it can accelerate the convergence of the training model and makes the network easier to train. (ii) It makes the different batches of training data keep uniform distribution and improves the performance of the network. (iii) It has low sensitivity for initialisation.

To the best of our knowledge, although increasing the depth of network can improve the performance for image denoising, deeper network may lead to the vanishing or exploding gradient problems. Residual learning is a good tool to solve this problem. It mainly adds the input (original images) and residual block (the output of several feature layers) as the input of the current layer to guarantee the performance. As shown in Fig. 1, we assume that $x$ and $f(x)$ represent the input and the output of stack several layers, respectively. The input of the next layer of the stack several layers is $f(x) + x$.

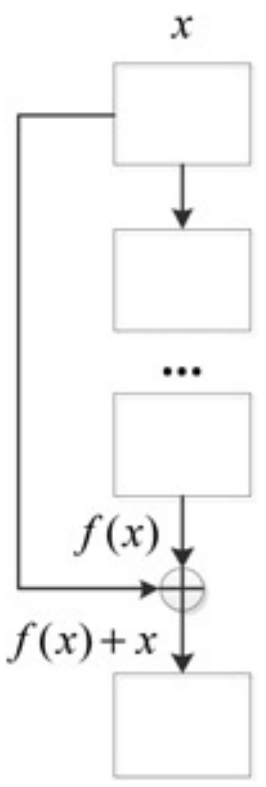

**Fig. 1** *Idea of residual learning mechanism*

### 2.2 Dilated convolution

As we know, more features can improve the performance for image processing [34–36]. Enlarging the receptive field in the CNN is very effective for extracting more features for image denoising [37]. There are two popular ways to enlarge the receptive field: (i) enlarging the width of the network (also referred to as increasing the filter size). (ii) Increasing the depth of the network. However, the first way may produce more parameters, which results in over fitting of the network. It also increases the computational cost. The second way may lead to vanishing/exploding gradients when the depth of the network is big. As a consequence, dilated convolution is a good choice to balance the above ways. Dilated convolution uses a dilated filter with dilation factor $f$ to increase the obtained information. That is, a dilated filer can be expressed as a filter with size $(2f + 1)(2f + 1)$. For example, when is 1, the receptive field of the first layer is 3. The receptive fields of the other layers are 5, 7, 9,…, respectively. In addition, combining the dilated filter and the convolutional kernel of $3 \times 3$ is very popular for image processing [29]. For more details on dilated convolution, please refer to [29].

## 3 Proposed denoising method

### 3.1 Network architecture

According to the previous research, we know that the denoising method can be expressed as $y = x + \mu$. In this paper, the objective function of learning $f(y)$ is as follows:

$$l(p) = \frac{1}{N}\sum_{j=1}^{N}\left\| f\left(y_j; p\right) - \left(y_j - x_j\right)\right\|^2 \quad (1)$$

Formula (1) is the objective function to train the denoising model, where $p$ represents the parameters, $y_j$ represents the $j$th noisy image patch and $x_j$ represents the $j$th label image patch. Specifically, the image patches can reduce the computational cost

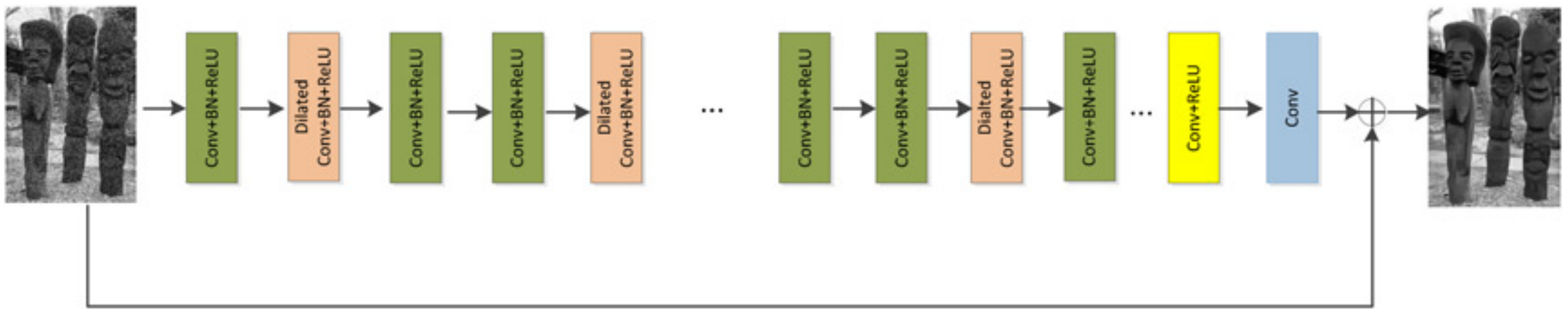

**Fig. 2** *Architecture of ECNDNet*

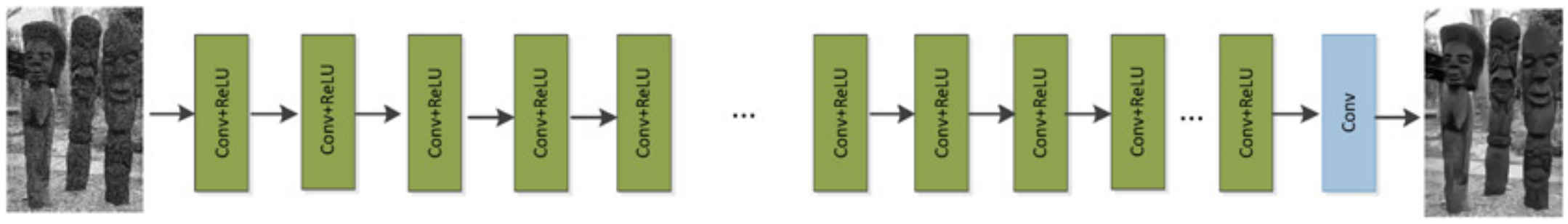

**Fig. 3** *Architecture of CRNet*

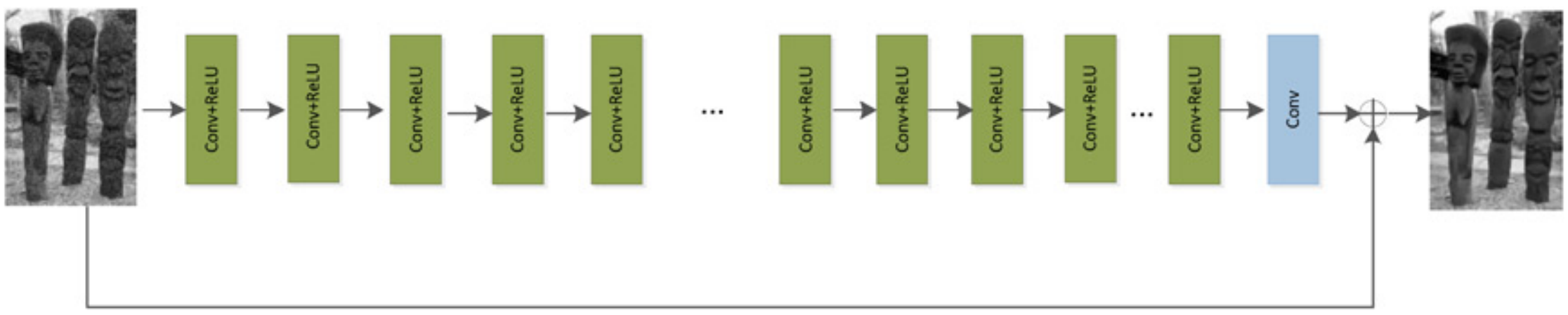

**Fig. 4** *Architecture of CRRNet*

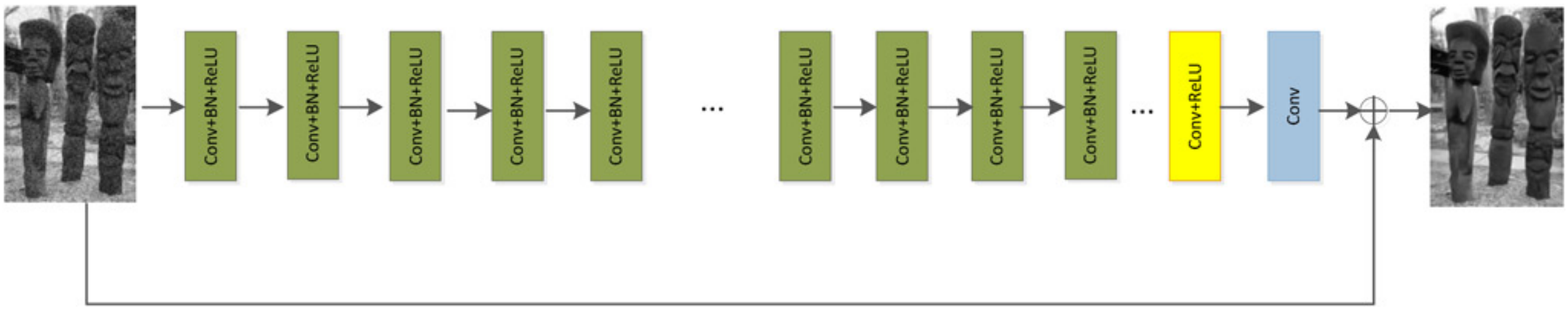

**Fig. 5** *Architecture of CRRBNet*

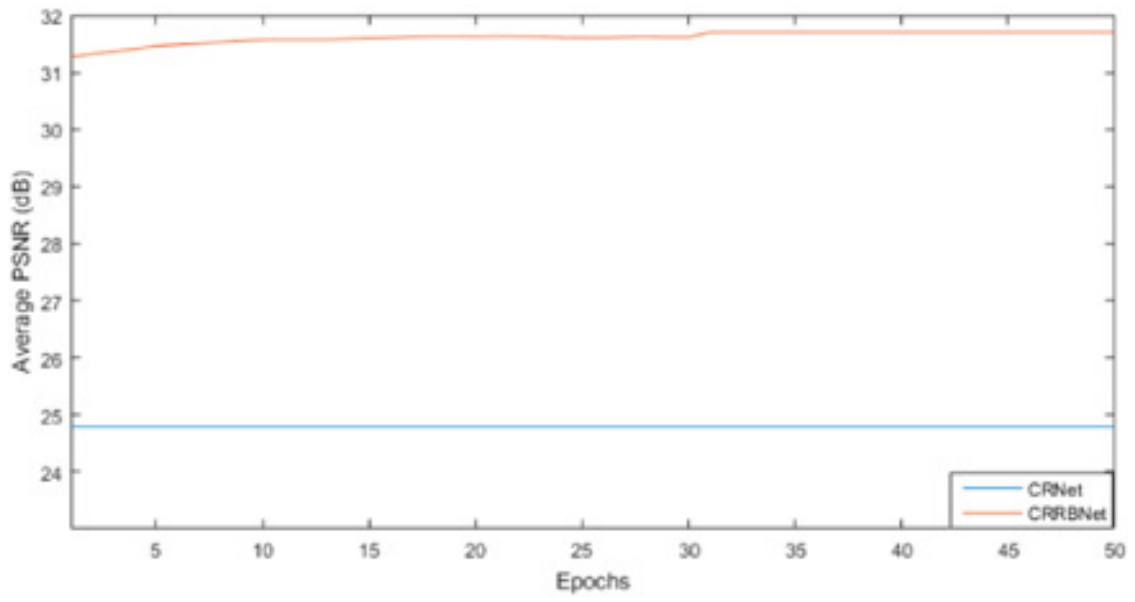

**Fig. 6** *Guassian denoising results of CRNet and CRRBNet on BSD68 is shown. CRNet only has convolution and ReLU. CRRBNet includes BN and ReLU and residual learning. They are trained with* $\sigma = 15$

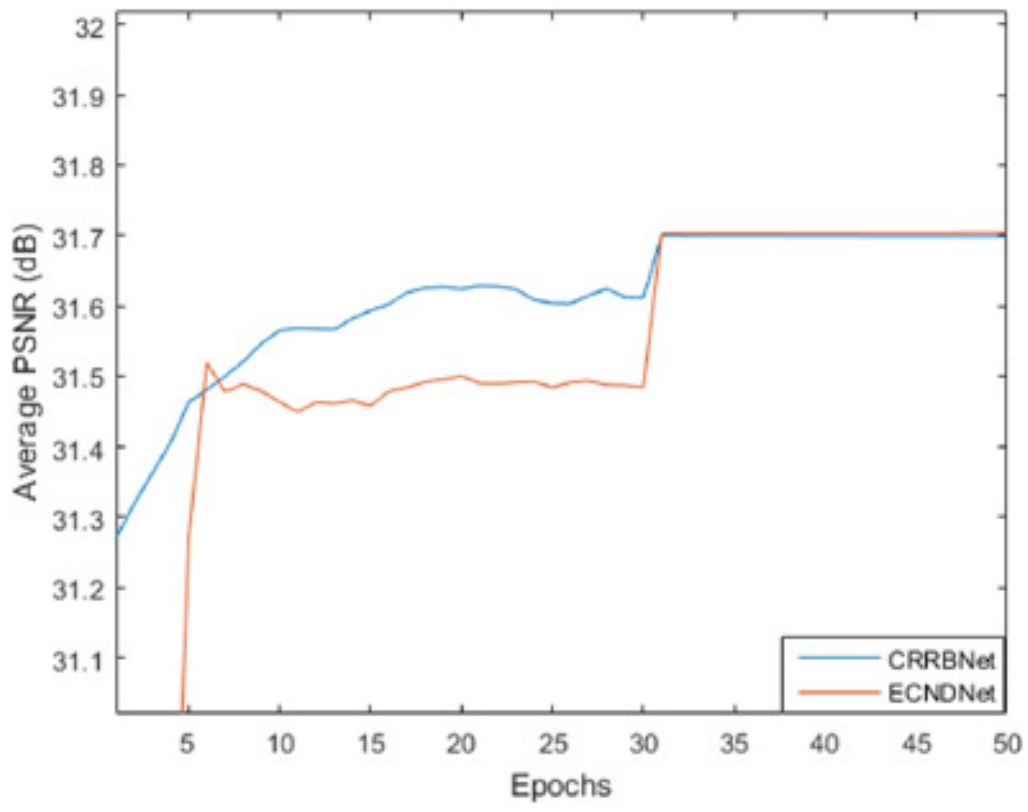

**Fig. 7** *Guassian denoising results of CRRBNet and ECNDNet on BSD68 is shown. CRRBNet has BN and ReLU and residual learning. ECNDNet includes BN, ReLU, residual learning and dilated convolution. They are trained with* $\sigma = 15$

and learn more features [38]. Thus, we divided the image into patches is reasonable for image denoising. In addition, very deep architecture is another non-ignorable factor which can result in vanishing or exploding gradient problems. As a result of these concerns, we proposed a novel network called ECNDNet. ECNDNet consists of dilated convolution, residual learning, BN, convolution (Conv) and ReLU. We empirically find that sets the dilated convolution to the 2nd, 5th, 9th and 12th layers can not only increase the captured information, but also reduces the computational cost than that of each layer with dilated convolution. Moreover, the use of BN and residual learning makes this network more effective for image denoising. The architecture of the designed network is shown in Fig. 2. Also the depth of the proposed network is 17. It has four types in this network: Conv, ReLU, BN and dilated Conv. Specifically, they are convolution, rectified linear units, BN and dilated convolution, respectively. The 1st and 16th layers are Conv + ReLU. The 2nd, 5th, 9th and 12th layers are dilated Conv + BN + ReLU. Specifically, the dilated factor is important to enlarge the receptive field for dilated convolution. Here we use dilated factor of 2 and the receptive fields of all 17 are 3, 7, 9, 11, 15, 17, 19, 21, 25, 27, 29, 33, 35, 37, 39, 41 and 43, respectively. It can map the context features from $3 \times 3$ to $43 \times 43$. The final layer is Conv. The other layers are Conv + BN + ReLU. The size of the convolutional kernels is $128 \times 1 \times 40 \times 40$ for the first and the last layers, respectively. The size of other convolutional kernels is $128 \times 64 \times 40 \times 40$.

The merits of the proposed method have three-fold: (i) it uses 17 layers network and residual learning to prevent the problems of vanishing or exploding gradients. (ii) It uses BN technique to accelerate convergence and make the network easier to train. (iii) It uses dilated convolutions to enhance the performance of the designed network and reduce the computational cost.

### 3.2 Discussion

The proposed method relies on residual learning, BN and dilated convolution, they are complementary for image denoising. In this part, we will prove the effectiveness of these methods for

**Table 1** Average PSNR (dB) results from different methods on BSD 68

| Methods | BM3D | WNNM | EPLL | MLP | CSF | TNRD | IRCNN | ECNDN |
|---|---|---|---|---|---|---|---|---|
| $\sigma = 15$ | 31.07 | 31.37 | 31.21 | — | 31.24 | 31.42 | 31.63 | 31.71 |
| $\sigma = 25$ | 28.57 | 28.83 | 28.68 | 28.96 | 28.74 | 28.92 | 29.15 | 29.22 |
| $\sigma = 50$ | 25.62 | 25.87 | 25.67 | 26.03 | — | 25.97 | 26.19 | 26.23 |

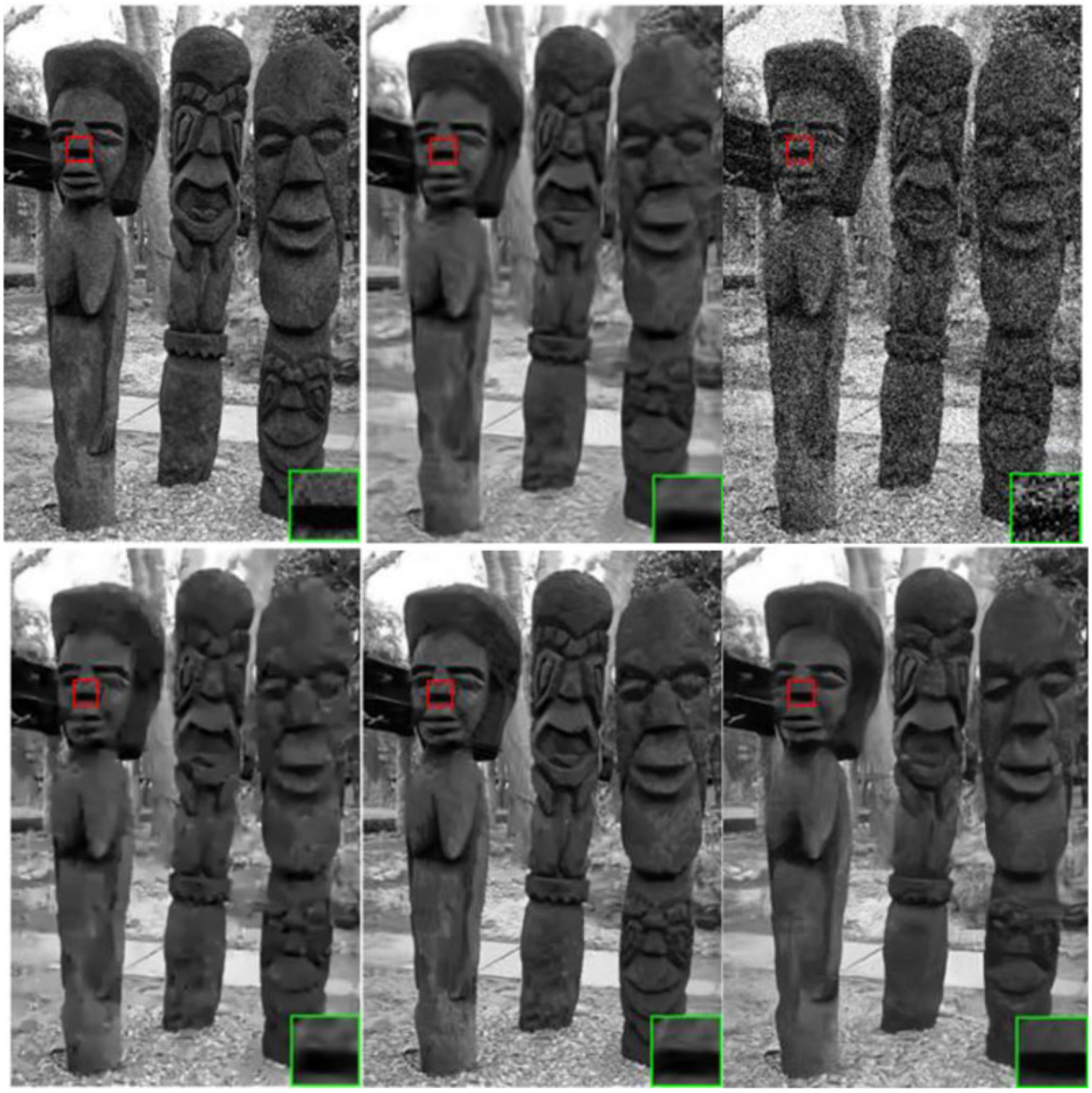

**Fig. 8** *Denoising results of one grey image from BSD68 with* $\sigma = 50$

*a* Original image
*b* BM3D/22.56 dB
*c* Noisy/15.07 dB
*d* WNNM/22.83 dB
*e* EPLL/22.81 dB
*f* ECNDNet/23.19 dB

**Table 2** Average PSNR (dB) results of different methods on widely used 12 images with noise levels 15, 25 and 50

| Images | C.man | House | Peppers | Starfish | Monarch | Airplane | Parrot | Lena | Barbara | Boat | Man | Couple | Average |
|---|---|---|---|---|---|---|---|---|---|---|---|---|---|
| noise level | | | | | | | $\sigma = 15$ | | | | | | |
| BM3D [6] | 31.91 | 34.93 | 32.69 | 31.14 | 31.85 | 31.07 | 31.37 | 34.26 | 33.10 | 32.13 | 31.92 | 32.10 | 32.37 |
| WNNM [13] | 32.17 | 35.13 | 32.99 | 31.82 | 32.71 | 31.39 | 31.62 | 34.27 | 33.60 | 32.27 | 32.11 | 32.17 | 32.70 |
| EPLL [38] | 31.85 | 34.17 | 32.64 | 31.13 | 32.10 | 31.19 | 31.42 | 33.92 | 31.38 | 31.93 | 32.00 | 31.93 | 32.14 |
| CSF [44] | 31.95 | 34.39 | 32.85 | 31.55 | 32.33 | 31.33 | 31.37 | 34.06 | 31.92 | 32.01 | 32.08 | 31.98 | 32.32 |
| TNRD [42] | 32.19 | 34.53 | 33.04 | 31.75 | 32.56 | 31.46 | 31.63 | 34.24 | 32.13 | 32.14 | 32.23 | 32.11 | 32.50 |
| IRCNN [17] | 32.55 | 34.89 | 33.31 | 32.02 | 32.82 | 31.70 | 31.84 | 34.53 | 32.43 | 32.34 | 32.40 | 32.40 | 32.77 |
| ECNDNet | 32.56 | 34.97 | 33.25 | 32.17 | 33.11 | 31.70 | 31.82 | 34.52 | 32.41 | 32.37 | 32.39 | 32.39 | 32.81 |
| noise level | | | | | | | $\sigma = 25$ | | | | | | |
| BM3D [6] | 29.45 | 32.85 | 30.16 | 28.56 | 29.25 | 28.42 | 28.93 | 32.07 | 30.71 | 29.90 | 29.61 | 29.71 | 29.97 |
| WNNM [13] | 29.64 | 33.22 | 30.42 | 29.03 | 29.84 | 28.69 | 29.15 | 32.24 | 31.24 | 30.03 | 29.76 | 29.82 | 30.26 |
| EPLL [38] | 29.26 | 32.17 | 30.17 | 28.51 | 29.39 | 28.61 | 28.95 | 31.73 | 28.61 | 29.74 | 29.66 | 29.53 | 29.69 |
| MLP [45] | 29.61 | 32.56 | 30.30 | 28.82 | 29.61 | 28.82 | 29.25 | 32.25 | 29.54 | 29.97 | 29.88 | 29.73 | 30.03 |
| CSF [44] | 29.48 | 32.39 | 30.32 | 28.80 | 29.62 | 28.72 | 28.90 | 31.79 | 29.03 | 29.76 | 29.71 | 29.53 | 29.84 |
| TNRD [42] | 29.72 | 32.53 | 30.57 | 29.02 | 29.85 | 28.88 | 29.18 | 32.00 | 29.41 | 29.91 | 29.87 | 29.71 | 30.06 |
| IRCNN [17] | 30.08 | 33.06 | 30.88 | 29.27 | 30.09 | 29.12 | 29.47 | 32.43 | 29.92 | 30.17 | 30.04 | 30.08 | 30.38 |
| ECNDNet | 30.11 | 33.08 | 30.85 | 29.43 | 30.30 | 29.07 | 29.38 | 32.38 | 29.84 | 30.14 | 30.03 | 30.03 | 30.39 |
| noise level | | | | | | | $\sigma = 50$ | | | | | | |
| BM3D [6] | 26.13 | 29.69 | 26.68 | 25.04 | 25.82 | 25.10 | 25.90 | 29.05 | 27.22 | 26.78 | 26.81 | 26.46 | 26.72 |
| WNNM [13] | 26.45 | 30.33 | 26.95 | 25.44 | 26.32 | 25.42 | 26.14 | 29.25 | 27.79 | 26.97 | 26.94 | 26.64 | 27.05 |
| EPLL [38] | 26.10 | 29.12 | 26.80 | 25.12 | 25.94 | 25.31 | 25.95 | 28.68 | 24.83 | 26.74 | 26.79 | 26.30 | 26.47 |
| MLP [45] | 26.37 | 29.64 | 26.68 | 25.43 | 26.26 | 25.56 | 26.12 | 29.32 | 25.24 | 27.03 | 27.06 | 26.67 | 26.78 |
| TNRD [42] | 26.62 | 29.48 | 27.10 | 25.42 | 26.31 | 25.59 | 26.16 | 28.93 | 25.70 | 26.94 | 26.98 | 26.50 | 26.81 |
| IRCNN [17] | 26.88 | 29.96 | 27.33 | 25.57 | 26.61 | 25.89 | 26.55 | 29.40 | 26.24 | 27.17 | 27.17 | 26.88 | 27.14 |
| ECNDNet | 27.07 | 30.12 | 27.30 | 25.72 | 26.82 | 25.79 | 26.32 | 29.29 | 26.26 | 27.16 | 27.11 | 26.84 | 27.15 |

image-denoising. Here CRNet, CRRNet and CRRBNet have the same the number of network layers, convolutional kernel size and initial parameters. Specifically, CRNet consists of Conv and ReLU as shown in Fig. 3, where Conv and ReLU denote the convolution and rectified linear units, respectively. CRRNet consists of Conv, ReLU and residual learning technique as shown in Fig. 4. Here Figs. 1–4 are the schematic diagrams, in this paper CRRBNet consists of Conv, ReLU, residual learning and BN as shown in Fig. 5. First, we illustrate the peak signal-to-noise ratio (PSNR) of every training epoch for CRNet and CRRBNet. From Fig. 6, we know that the combination of BN and residual learning is effective for image denoising. Then, we prove that the dilated convolution is useful for image denoising as shown in Fig. 7.

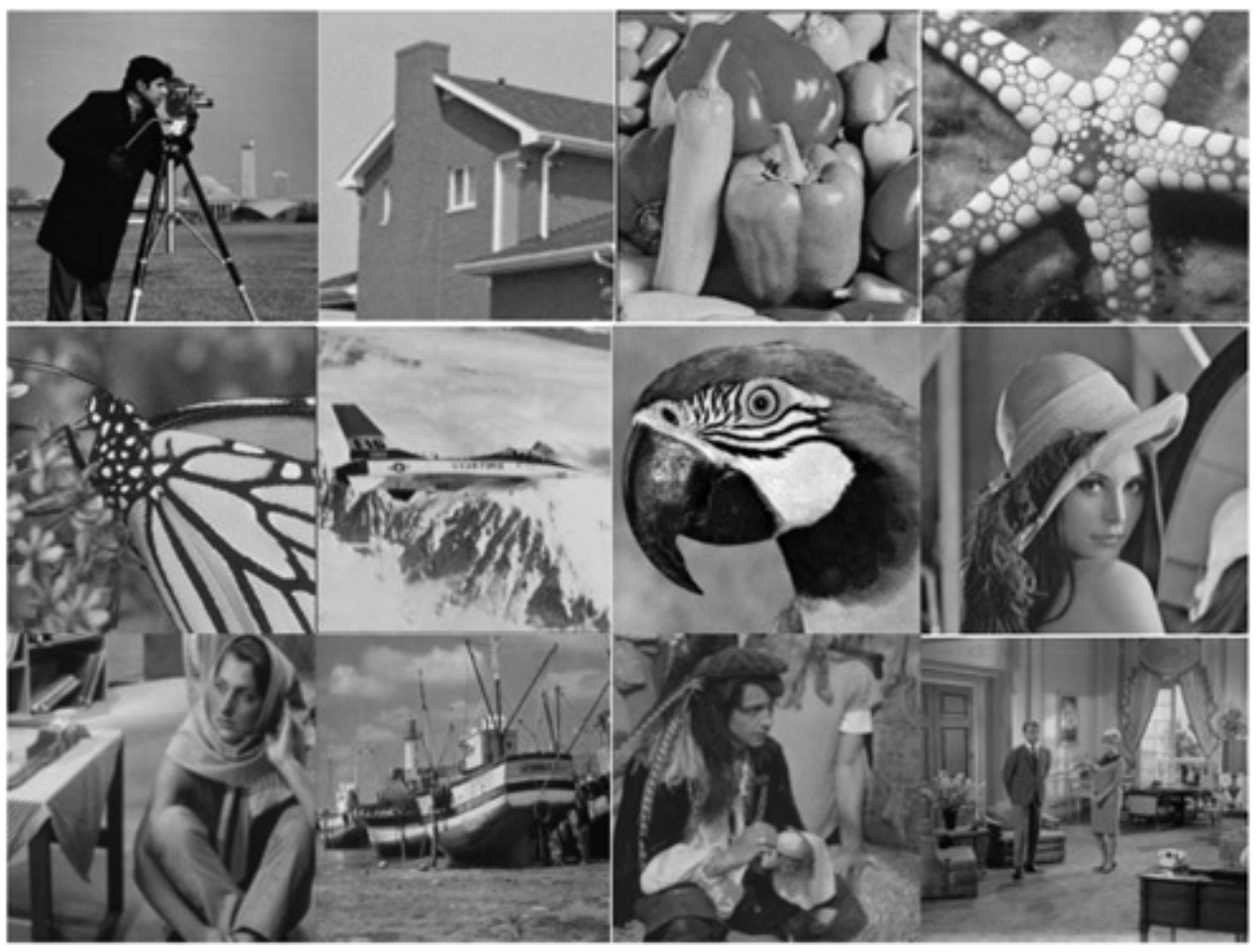

**Fig. 9** *Widely used 12 images*

## 4 Experimental results

### 4.1 Experimental setting

We design a 17-layer network called ECNDNet. Its depth is the same as denoising CNN (DnCNN). Its loss function (also referred to as objective function) is shown as in (1). We choose Adam [39] to optimise the converge model. The initial parameters are set as follows: (i) learning rate, beta_1, beta_2 and epsilon are $1\times10^{-3}$, 0.9, 0.999 and $1\times10^{-8}$, respectively. (ii) The initial weights are set as shown in [40]. (iii) The number of batches is 128. (iv) The number of epochs is 180 for the trained model. In addition, the learning rates of the 180 epochs are $1\times10^{-3}$ to $1\times10^{-8}$.

We choose PyTorch tool [41] to train the denoising model in this paper. All the experiments are implemented in the environment of Ubuntu 16.04 and python 2.7 and run on PC with Intel Core i7 7800X CPU, RAM 16G and a Nvidia GeForce GTX 1080 Ti GPU. The types of Nivdia CUDA and cuDNN are 9.0 and 7.5, respectively.

### 4.2 ECNDNet for grey image denoising

We choose 400 images [42] with size of $180 \times 180$ for Gaussian denoising. The format of training images is '.png'. According to the IRCNN [17] and fast and flexible denoising network (FFDNet) [15], we use BSD68 [43] and Set 12 to test the denoising model. In addition, we use popular methods such as BM3D [6], WNNM [13], expected patch log likelihood (EPLL) [38], cascade of shrinkage fields (CSF) [44], trainable nonlinear reaction diffusion (TNRD) [42], IRCNN [17] and multi-layer perceptron [45] to verify the performance of gray image denoising. To test the robustness of our proposed method for low-level and high-level noise, we choose $\sigma = 15$, $\sigma = 25$ and $\sigma = 50$ to conduct comparative experiments. For example, the PSNR of our proposed method is 31.71 dB higher than that of the state-of-the-art method such as IRCNN as shown in Table 1 ($\sigma = 15$). Besides, the

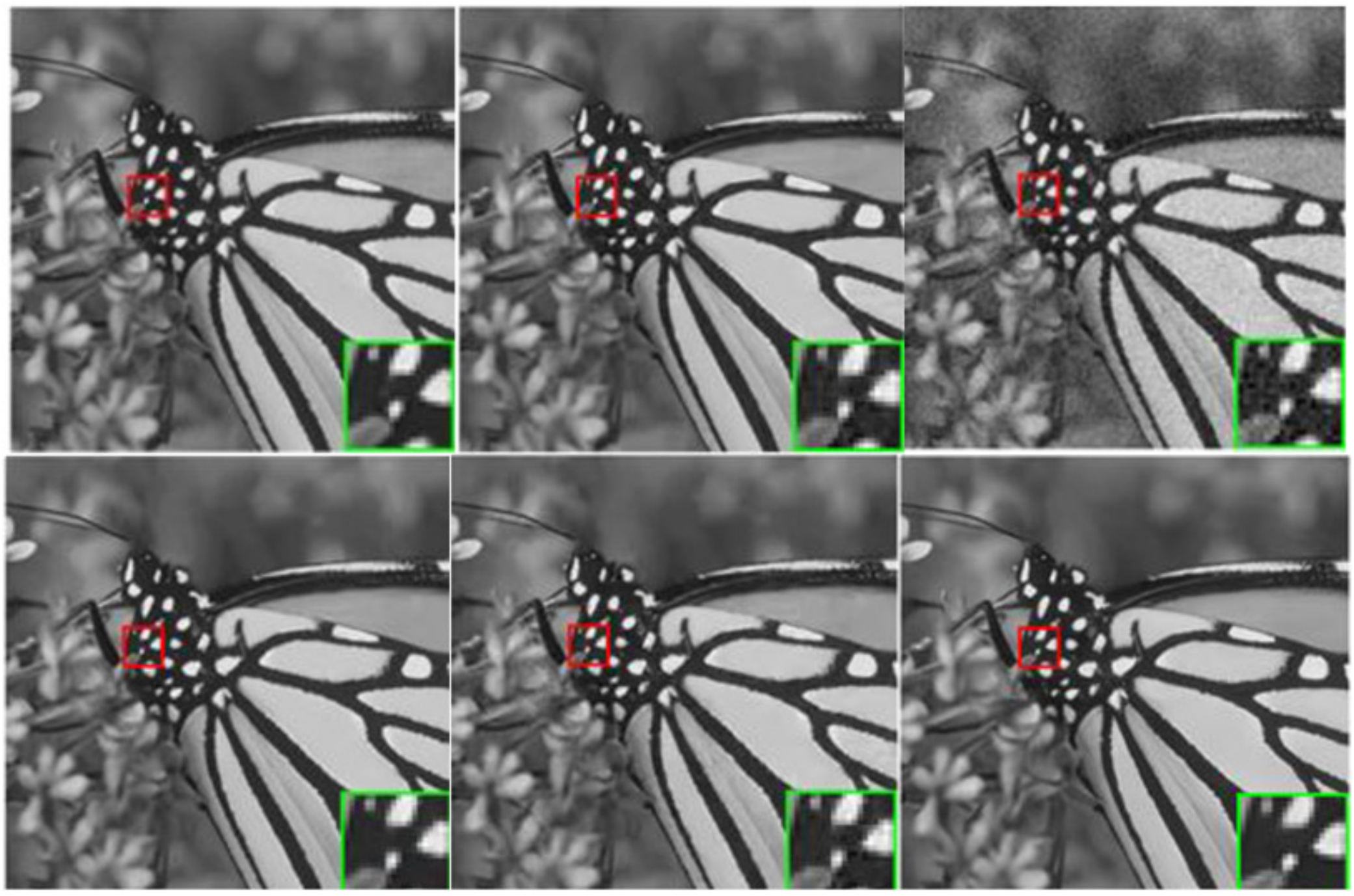

**Fig. 10** *Denoising results of one grey image with* $\sigma = 15$

*a* Original image
*b* BM3D/31.85 dB
*c* Noisy/24.59 dB
*d* WNNM/32.71 dB
*e* EPLL/32.10 dB
*f* ECNDNet/33.11 dB

**Table 3** Run time of different methods on image of size 256 × 256, 512 × 512 and 1024 × 1024 with noise level 25

| Methods | Device | 256 × 256 | 512 × 512 | 1024 × 1024 |
|---|---|---|---|---|
| | | Gray | Gray | Gray |
| BM3D | CPU | 0.65 | 2.85 | 11.89 |
| WNNM | CPU | 203.1 | 773.2 | 2536.4 |
| EPLL | CPU | 25.4 | 45.5 | 422.1 |
| MLP | CPU | 1.42 | 5.51 | 19.4 |
| CSF | CPU/GPU | 2.11/— | 5.67/0.92 | 40.8/1.72 |
| TNRD | CPU/GPU | 0.45/0.010 | 1.33/0.032 | 4.61/0.116 |
| DnCNN-s | GPU | 0.008 | 0.068 | 0.154 |
| ECNDNet | GPU | 0.012 | 0.079 | 0.205 |

best and second best performance are shown in italic and bold, respectively.

We use Fig. 8 to vividly show the performance of our method and other comparative methods with $\sigma = 50$ on BSD68 dataset. To show the performance of our proposed method for the images of different categories, we validate it using the Set12 dataset.

From Table 2, it is known that our proposed method has good performance for each category image (Fig. 9). For example, the average PSNR of our method is 30.39 dB higher than that of BM3D when noise level is 25. Specifically, the best PSNR is marked in italic and the second PSNR is marked in bold as shown in Table 2. The detailed results of the comparative experiments are shown in [3, 15, 17]. Fig. 10 shows the denoising performance of different methods of an image.

## 5 Run time

PSNR and run time of processing an image are two important factors of image denoising. The performance of the proposed method has been proved in Section 4.2. The run time of processing an image is tested for gray image denoising as follows. We utilise noisy image sizes of 256 × 256, 512 × 512 and 1024 × 1024 with $\sigma = 50$ to test the speed of different methods for an image. Specifically, we use PyTorch to test run time of DnCNN-s and ECNDNet. From Table 3, we know that our ECNDNet is competitive with popular methods such as BM3D, WNNM, EPLL, CSF, TNRD and DnCNN-s in run time. In summary, our proposed method is robust for image denoising.

## 6 Conclusion

In this paper, a deep CNN called ECNDNet is proposed to solve the image denoising problem.

Specifically, BN, residual learning and dilated convolution are used to enhance network performance. BN can deal with internal covariate shift problem and makes the network easier to train. Residual learning technique can address the problem of vanishing or exploding gradients. It is used to obtain clean images from noisy images and residual images. Dilated convolution can extract more context information and reduce the computational cost. In addition, BN, residual learning and dilated convolution are complement for image denoising. Extensive experiments show that ECNDNet is more effective than the popular denoisng methods such as IRCNN. In the future, we will combine model base-optimisation and discriminative learning methods to remove the noise from real noisy images.

## 7 Acknowledgments

This paper was supported in part by the Guangdong Province high-level personnel of special support program under grant no. 2016TX03X164, in part by the Shenzhen Municipal Science and Technology Innovation Council under grant no. JCYJ20170811155725434.